\documentclass[10pt, a4paper]{article}
\usepackage{lrec2022} 
\usepackage{multibib}
\newcites{languageresource}{Language Resources}
\usepackage{graphicx}
\usepackage{tabularx}
\usepackage{multirow}
\usepackage{soul}
\usepackage{scalefnt}
\usepackage{titlesec}
\titleformat{\section}{\normalfont\large\bfseries\center}{\thesection.}{1em}{}
\titleformat{\subsection}{\normalfont\SmallTitleFont\bfseries\raggedright}{\thesubsection.}{1em}{}
\titleformat{\subsubsection}{\normalfont\normalsize\bfseries\raggedright}{\thesubsubsection.}{1em}{}
\renewcommand\thesection{\arabic{section}}
\renewcommand\thesubsection{\thesection.\arabic{subsection}}
\renewcommand\thesubsubsection{\thesubsection.\arabic{subsubsection}}
\usepackage{epstopdf}
\usepackage[utf8]{inputenc}
\usepackage{hyperref}
\usepackage{xstring}
\usepackage{color}

\title{HateBR: A Large Expert Annotated Corpus of Brazilian Instagram Comments for Offensive Language and Hate Speech Detection}

\name{\large Francielle Vargas*†, Isabelle Carvalho*, Fabiana Góes* \\ \textbf{\large Thiago A.S. Pardo*, Fabrício Benevenuto†}} 

\address{*Institute of Mathematical and Computer Sciences, University of São Paulo, Brazil \\  †Computer Science Department, Federal University of Minas Gerais, Brazil\\
 \{francielleavargas,isabelle.carvalho,fabianagoes\}@usp.br, 
 taspardo@icmc.usp.br, fabricio@dcc.ufmg.br\\}

\abstract{
Due to the severity of the social media offensive and hateful comments in Brazil, and the lack of research in Portuguese, this paper provides the first large-scale expert annotated corpus of Brazilian Instagram comments for hate speech and offensive language detection. The HateBR corpus was collected from the comment section of Brazilian politicians' accounts on Instagram and manually annotated by specialists, reaching a high inter-annotator agreement. The corpus consists of 7,000 documents annotated according to three different layers: a binary classification (offensive versus non-offensive comments), offensiveness-level classification (highly, moderately, and slightly offensive), and nine hate speech groups (xenophobia, racism, homophobia, sexism, religious intolerance, partyism, apology for the dictatorship, antisemitism, and fatphobia). We also implemented baseline experiments for offensive language and hate speech detection and compared them with a literature baseline. Results show that the baseline experiments on our corpus outperform the current state-of-the-art for the Portuguese language. 
 \\ \newline \Keywords{abusive language detection, offensive language and hate speech, corpus annotation, natural language processing} }

\begin{document}

\maketitleabstract

\section{Introduction}
Abusive language detection has attracted interest from different institutions and has become an important research topic \cite{Polettoetall2021,pitenisetal2020offensive,coltekin2020corpus,guestetal2021expert}. While this challenging endeavour is undoubtedly a relevant research line, also has its implications for society concerning race, gender, religion, and origin. Furthermore, automated methods for hateful and offensive comments detection may bolster web security in revealing individuals harboring malicious intentions towards specific groups \cite{gaoetal2017recognizing}.

In Brazil, hate speech is prohibited, nevertheless the regulation is not effective due to the difficulty of identifying, quantifying and classifying this kind of online content. The data on hate crimes in Brazil is very worrisome: during the 2018 year's election period, denunciations with xenophobia content had an increase of 2,369\%; apology and public incitement to violence and crimes against life, 630\%; neo-nazism, 548\%; homophobia, 350\%; racism, 218\%; and religious intolerance, 145\% \footnote{\url{https://www.bbc.com/portuguese/brasil-46146756}}.

The state-of-the-art has focused on different tasks, such as automatically detecting hate speech groups, for example, racism  \cite{hasanuzzamanetal2017demographic}, antisemitism \cite{SavvasetAll2020}, religious intolerance \cite{ghoshchowdhuryetal2019arhnet}, misogyny and sexism \cite{guestetal2021expert,jhamamidi2017compliment}, and cyberbullying  \cite{safi-samghabadi-etal-2020-detecting}; filtering pages with hate and violence \cite{LiuF15a}; offensive language detection \cite{zampierietal2019,steimeletal2019investigating}; and toxicity \cite{leite-etal-2020-toxic,samuelEtall2020}. \newcite{schmidtwiegand2017survey} present a comprehensive survey of Natural Language Processing (NLP) techniques applied to hate speech detection, and  \newcite{Polettoetall2021} describe resources and benchmark corpora for hate speech detection. Multilingual hate speech detection is studied by \newcite{ranasinghe-zampieri-2020-multilingual,steimeletal2019investigating,basile-etal-2019-semeval}. 

Due to the relevance of this topic and the severity of online hate speech context in Brazil, the proposition of a reliable annotated corpus is fundamental to carry out experiments and to build automatic systems for abusive language detection. Nevertheless, the annotation process this kind of content is intrinsically challenging, bearing in mind that what is considered offensive is influenced by pragmatic (contextual) factors, and people may have different perspectives on an offense. On account of that, \newcite{Polettoetall2021} claim that authors in the field have discussed aspects related to the implications of an annotation process for offensive language and hate speech phenomena, which inspired a multi-layer annotation scheme \cite{zampierietal2019}, target-aware annotation \cite{basile-etal-2019-semeval}, and the implicit-explicit distinction in the annotation \cite{caselli-etal-2020-feel}. Corroborating those authors, we claim that, as being particularly challenging the offensive language and hate speech detection, a well-defined annotation schema has a considerable impact among the consistency and quality of the data, and the performance of the derived machine learning classifiers.

In this paper, we provide the first large-scale expert annotated corpus of Brazilian Instagram comments for abusive language detection in Brazilian Portuguese. We assume that hate speech and offensive language are types of abusive language. The HateBR corpus was collected from different accounts of Brazilian politicians from  Instagram social media. The political context was chosen due to the identification of a wide variety of serious offensive and hateful attacks against different groups. The entire annotation schema was proposed and annotated by different specialists: a linguist, a hate speech expert, NLP and machine learning researchers, and handled by accurate guidelines and training steps, in order to ensure the same understanding of the tasks, and towards minimizing bias. Furthermore, baseline experiments were implemented, whose results (85\% of F1-score) overcame the current state-of-the-art for the Portuguese language. More precisely, the main contributions of this paper are: 

\begin{itemize}
    \item The first large-scale expert annotated corpus for offensive language and hate speech on the web and social media in Brazilian Portuguese. The corpus titled ``HateBR'' consists of 7,000 Instagram comments annotated in three different layers (offensive versus non-offensive; offensive comments sorted into offensiveness-levels such as highly, moderately, and slightly; and nine hate speech groups: xenophobia, racism, homophobia, sexism, religious intolerance, partyism, an apology to dictatorship, antisemitism, and fatphobia).
    
    \item A new expert annotation schema for hate speech and offensive language detection, which is divided into three layers: offensive classification, offensiveness-classification and hate speech classification.
    
\end{itemize}

In what follows, we briefly introduce the main related work. Section \ref{sec:corpus_development} describes the HateBR corpus development, as well as the proposed annotation schema and its evaluation. In Sections \ref{sec:hatebrstatistics} and
\ref{sec:hatebr_ex}, HateBR corpus statistics and experiments are presented. At last, final remarks are discussed in Section \ref{sec:final_remarks}.

\section{Related Work}
\label{sec:related_work}
Most of hate speech and offensive language corpora are proposed for the English language \cite{zampierietal2019,FersiniRA18,DavidsonWMW17,gaohuang2017detecting,jhamamidi2017compliment,golbecketall2017}. For the French language, a corpus of Facebook and Twitter annotated data for Islamophobia, sexism, homophobia, religion intolerance and disability detection was also proposed \cite{chungetal2019conan,ousidhoumetal2019multilingual}. For the German language, a new anti-foreigner prejudice corpus was proposed. This corpus is composed of 5,836 Facebook posts and hierarchically annotated with slightly and explicitly/substantially offensive language according to six targets: foreigners, government, press, community, other, and unknown \cite{BretschneideretRalf2017german}. For the Greek language, an annotated corpus of Twitter and Gazeta posts for offensive content detection is also available \cite{pitenisetal2020offensive,pavlopoulosetal2017deep}. For the Slovene and Croatian languages, a large-scale corpus composed of 17,000,000 posts, composed of 2\% of abusive language on a leading media company website was built \cite{ljubesicetal2018datasets}. For the Arabic language, there is a corpus of 6,136 twitter posts, which is annotated according to religion intolerance subcategories \cite{AlbadiEtAll2018}. For the Indonesian language, a hate speech annotated corpus from Twitter data was also proposed \cite{alfina2017hate}.

For the Portuguese language, a corpus composed of 5,668 tweets in European and Brazilian Portuguese, and automated methods using a hierarchy of hate to identify social groups of discrimination was proposed by \newcite{fortuna2019hierarchically}. They used pre-trained Glove word embeddings with 300 dimensions for feature extraction and a LSTM architecture proposed in \newcite{Badjatiyaetall2017}. The authors obtained 78\% of F1-score using cross-validation. Furthermore,  \newcite{fortuna-etal-2021-min} built a new specialized lexicon  specifically for European Portuguese, which, according to the authors, may be useful to detect a broader spectrum of content referring to minorities. Also, for Brazilian Portuguese, a corpus composed of 1,250 comments collected from G1 Brazilian online newspaper \footnote{\url{https://g1.globo.com/}} was proposed by \newcite{MoreiraAndPelle2017}. The authors report the annotation of a binary class: offensive and non-offensive comments and seven hate groups (racism, sexism, homophobia, xenophobia, religious intolerance, and cursing). The authors evaluated a set of features based in n-grams and Information Gain  (InfoGain) algorithm \cite{witten2016data}. Classical machine learning methods such as Support Vector Machine (SVM) \cite{scholkopf2001learning} with linear kernel, and Multinomial Naive Bayes (NB) \cite{eyheramendy2003naive} were applied. The best model obtained 80\%  of F1-Score. 

\section{HateBR Corpus Development}
\label{sec:corpus_development}
In this section, we describe in detail the process for the building, annotation, and evaluation of the proposed corpus.

\subsection{Approach Overview}
\begin{figure}[!htbp]
    \centering
    \includegraphics[width=0.5\textwidth]{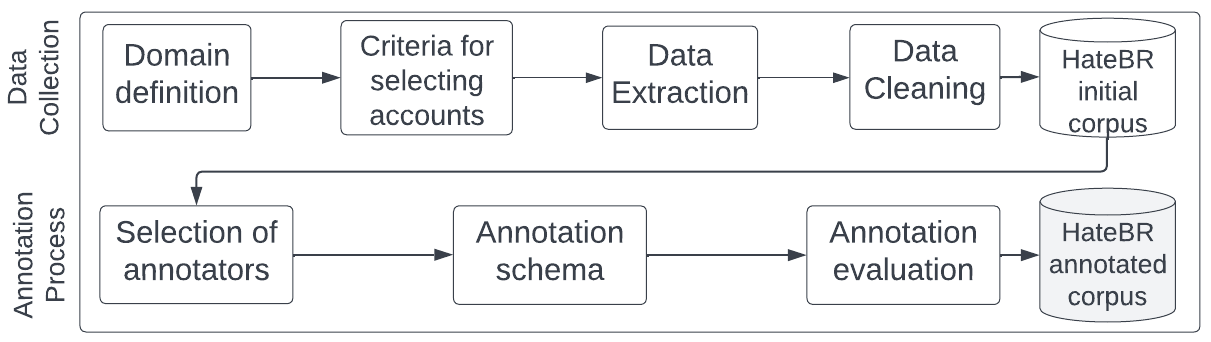}
    \caption{The proposed approach for the HateBR corpus construction.}
    \label{fig:ourapproach}
\end{figure}

The entire process for the corpus construction occurred for approximately six months, between August 2020 to January 2021. This project was performed by different specialists (e.g., a linguist, a hate speech expert, and NLP and machine learning researchers) and led by the linguist and hate speech expert in order to ensure the reliability and quality of the annotated data. Figure \ref{fig:ourapproach} exhibits an overview of the proposed approach for the construction of the HateBR corpus.

As shown by Figure \ref{fig:ourapproach}, in the first step - domain definition - the political domain was selected. In the second step - criteria for selection of accounts - the following criteria were defined: six different public accounts, being three liberal-party and three conservative-party accounts, from four women and two men. In the third step - data extraction - we implemented an Instagram API using the following parameters: post id, maximum number of 500 comments per post, and only public accounts were selected. Then, we extracted five hundred comments for each post published across six months from the second half of 2019 year were collected. For instance, five hundred comments were collected from the same account on an Instagram post published in August 2019. In the same settings, other five hundred comments were collected from a second post published in September 2019, and so on. In total, thirty posts were selected from six predefined Instagram accounts. Subsequently to the data extraction step, we proposed an approach for data cleaning. The data cleaning step basically consists of removing noise, such as links, characters without semantic value, and also comments that presented only emoticons, laughs (kkk, hahah, hshshs), or mentions (e.g., @namesomeone) without any textual content. Hashtags and emotions were kept. After these steps, we obtained the initial version of HateBR corpus without labels. 

For the corpus annotation, we defined a set of criteria for selection of annotators, such as higher levels of education (e.g., Ph.D. and Ph.D. candidate); only specialists  (e.g., linguists, hate speech experts and computer scientists); and diverse profiles, such as distinct political orientations and colors in order to minimize bias. Subsequently, we began the annotation process and proposed a new annotation schema, determining more precisely offensive language and hate speech classification. After all the previous steps are completed, the corpus was annotated using different levels of classification. The first level consists of a binary classification in offensive language versus non-offensive language; each of the 7,000 Instagram comments was annotated with an offensive (3,500 comments) or a non-offensive (3,500 comments) label. The second layer consists of offensiveness-level classification (highly, moderately, and weakly). Each of the 3,500 comments classified as offensive in the first layer was classified in offensiveness levels: highly offensive (778 comments), moderately offensive (1,044 comments), and slightly offensive (1,678 comments). Lastly, in the third layer, offensive comments that incited violence or hate against groups, based on specific characteristics (e.g., physical appearance, religion) received a label of hate speech (727 comments) considering nine identified hate speech groups (xenophobia, racism, homophobia, sexism, religious intolerance, partyism, apology for the dictatorship, antisemitism, and fatphobia). Still in the third layer, offensive comments that did not present violence or hate against groups received a label of no hate speech (2,773 comments). Finally, we evaluated the proposed annotation process using annotation agreement metrics, as Kappa \cite{mchugh2012interrater,sim2005kappa} and Fleiss \cite{fleiss1971measuring}, which obtained a high inter-annotator agreement for offensive language classification (75\% Kappa and 74\% Fleiss), and moderate for offensiveness-level classification (47\% Kappa and 46\% Fleiss).

\subsection{Data Collection}
Brazil occupies the third position in the worldwide ranking of Instagram's audience with 110 million active Brazilian users with an audience of 93 million users\footnote{\url{https://www.statista.com/}}. Taking into consideration that Instagram is a powerful platform for mass media, we automatically collected Instagram comments to build our corpus. Tables \ref{tab:data1} and \ref{tab:accounts} show the data collection statistics.

\begin{table}[!htbp]
\centering
\caption{Data collection statistics.} 
\label{tab:data1}
\scalefont{0.85}
\begin{tabular}{l|l}
\hline
\textbf{Data} & \textbf{Total} \\\hline
Amount of extracted comments  & 15,000 \\
Amount of removed comments   & 8,000 \\ 
\hline
Final corpus & 7,000 \\
\hline
\end{tabular}
\end{table}

\begin{table}[!htbp]
\centering
\caption{Accounts and posts information.} 
\label{tab:accounts}
\scalefont{0.85}
\begin{tabular}{l|l|l}
\hline
\textbf{Profile} & \textbf{Total} & \textbf{Description}  \\\hline
Gender      & 6 accounts & 4 women and 2 men  \\
Political   & 6 accounts & 3 liberals and 3 conservative \\
Posts       & 30 posts & 500 comments per post\\ \hline
\end{tabular}
\end{table}

As shown in Tables \ref{tab:data1} and \ref{tab:accounts}, corroborating our proposal of balancing the variables, such as gender and political party, we collected fifteen thousand comments from six public Instagram accounts of Brazilian politicians divided in three politicians from the liberal party, and three politicians from the conservative party, being four women and two men. We decided to select the most popular posts for each account during the second half of 2019, being five posts for each account and five hundred comments for each post. Thereafter, we removed eight thousand comments that presented only emoticons, laughs or mentions. In addition, labeled comments that were surplus aiming at balancing the classes of binary classification were also removed. Therefore, in these eight thousand removed items, there are both noises and surplus labeled comments. 

\subsection{Annotation Process}
\label{sec:annotation_process}
A detailed description of our annotation process approach is presented in this Section.

\subsubsection{Selection of Annotators}
The first step of the annotation process consists of selection of annotators. Due to the degree of complexity of the offensive language and hate speech detection tasks,  mainly because it involves a highly politicized domain, we decided to select only specialists at higher levels of education. In addition, towards minimizing bias and their negative impact on the results, we diversify the annotators' profile, as shown in Table \ref{tab:annotators}.

\begin{table}[!htbp]
\centering
\caption{Annotators' profile.} 
\label{tab:annotators}
\scalefont{0.85}
\begin{tabular}{l|l}
\hline
\textbf{Profile}       & \textbf{Description}  \\
\hline
Education & PhD or PhD candidate  \\
Gender & feminine  \\
Political & liberal and conservative  \\
Color & black and white \\
Brazilian region & north and southeast \\
\hline
\end{tabular}
\end{table}

As shown in Table \ref{tab:annotators}, the annotators are from North and Southeast Brazilian regions, and have at least a PhD study. Furthermore, they are white and black women, and are aligned with liberal or conservative parties.

\subsubsection{Annotation Schema}
\begin{figure*}[!htb]
    \centering
    \includegraphics[width=0.81\textwidth]{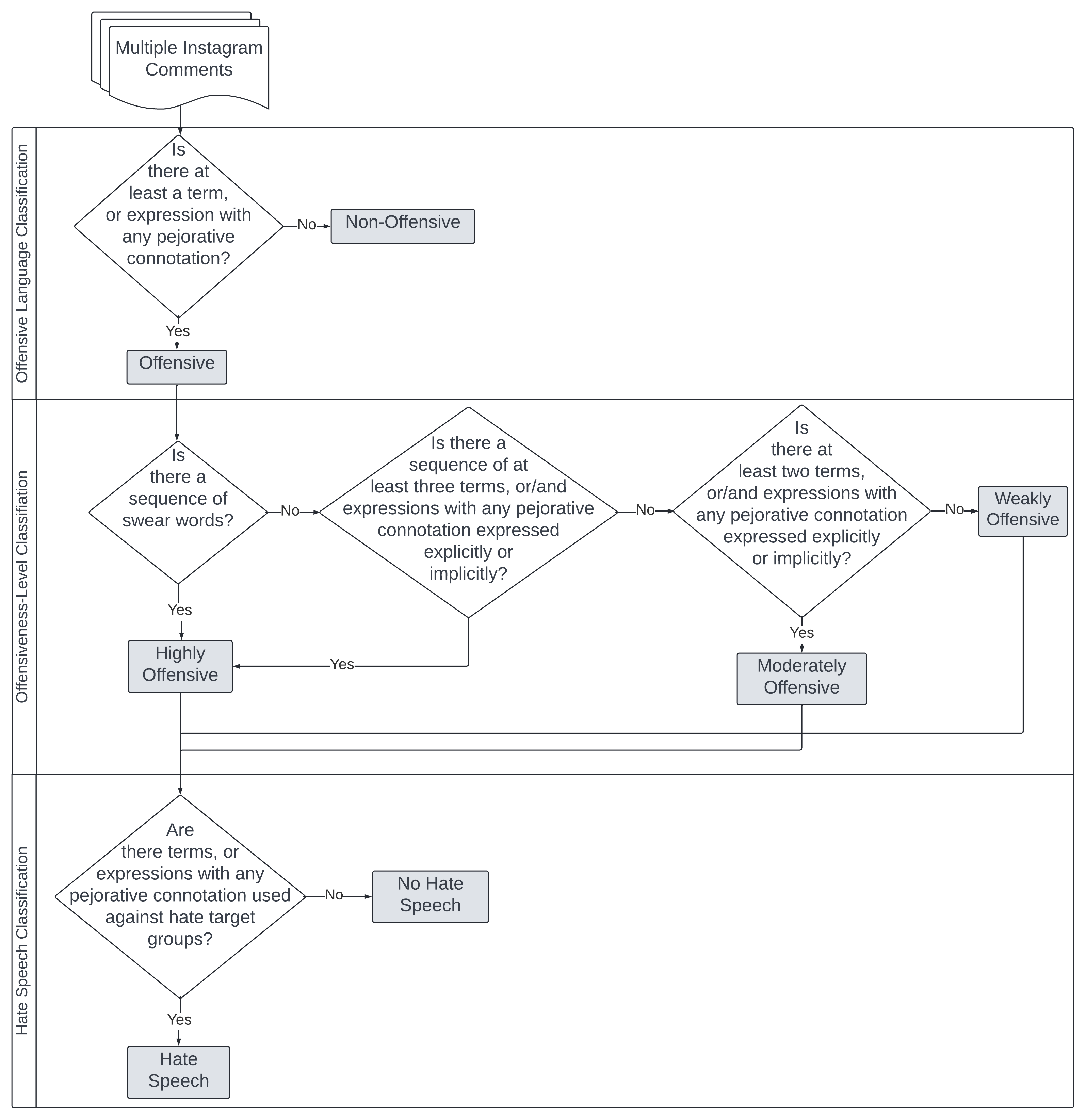}
    \caption{HateBR annotation schema.}
    \label{fig:schema}
\end{figure*}

Matter of ongoing debate, offensive language and hate speech detection tackles a conceptual difficulty distinguishing hateful and offensive expressions from expressions that merely denote dislike or disagreement \cite{Post2009}. In spite of the enormous difficulty of these tasks, this paper provides a new annotation schema for automatic detection of offensive language and hate speech in Brazilian Portuguese, as shown in Figure \ref{fig:schema}. In our annotation schema, we accurately discriminate each one of these definitions - offensive language and hate speech - which will be described in the following paragraphs.

According to \newcite{zampierietal2019}, offensive posts include insults, threats, and messages containing any form of untargeted profanity. Accordingly, in this paper we assume that offensive language consists of a kind of language containing terms or expressions with any pejorative connotation, including swear words\footnote{Swear words express the speaker’s emotional state tied to impoliteness and rudeness speech. They are a type of opinion that is highly confrontational, rude, or aggressive \cite{JayetJanschewitz2008,culpeperetall2017}}, which may be explicit or implicit. Furthermore, as defined by \newcite{FortunaAndNunes2018}, in this paper we assume that hate speech is a kind of language that attacks or diminishes, that incites violence or hate against groups, based on specific characteristics such as physical appearance, religion, or others, and it may occur with different linguistic styles, even in subtle forms or when humor is used. Therefore, hate speech is a type of offensive language used against groups target of discrimination (e.g., sexism, racism, homophobia). 

Table \ref{tab:odefinitions} shows examples of offensive language and hate speech, which may be explicit or implicit, extracted from the HateBR corpus. Note that \textbf{bold} indicates terms or expressions with explicit pejorative connotation, and \underline{underline} indicates ``clues'' of terms or expressions with implicit pejorative connotation. We also describe the terms originally written in Portuguese and their translation to English.

\begin{table}[!htb]
\caption{Examples of comments classified as offensive language and hate speech extracted from the HateBR corpus.}
\label{tab:odefinitions}
\centering
\scalefont{0.85}
\begin{tabular}{p{15mm}|p{24mm}|p{23mm}}
\hline
\textbf{Type} &  \textbf{Instagram Comments} & \textbf{Translation} \\
\hline
Offensive Language & Essa \textbf{besta humana} é o \textbf{câncer} do País, tem q \underline{voltar para jaula}, urgentemente! E viva o Presidente Bolsonaro. & This \textbf{human beast} is the \textbf{cancer} of the country, it has to \underline{go back to the cage}, urgently! And long live President Bolsonaro.\\
\hline
Non-Offensive Language & Quem falou isso pra vc deputada? O sergio moro ta aprovado pela maioria dos brasileiros. & Who said that to you deputy? Sergio Moro is approved by the majority of Brazilians.  \\
\hline
Hate Speech & \textbf{Vagabunda}. \textbf{Comunista}. \textbf{Mentirosa}. O povo chileno nao merece uma \textbf{desgraça} desta & \textbf{Bitch}. \textbf{Communist}. \textbf{Liar}. The people from Chile do not deserve such a \textbf{disgrace}. \\
\hline
No-Hate Speech & Pois é, deveria \underline{devolver o dinheiro} aos cofres públicos do Brasil. \textbf{Canalha}. & It should \underline{refund money} to the public Brazilian coffers. \textbf{Jerk}.\\
\hline
\end{tabular}
\end{table}

 As it is shown in Table \ref{tab:odefinitions}, there are explicit and implicit terms or expressions with pejorative connotation in offensive and hate speech comments. For example, in the comments classified as \textit{offensive language} and \textit{no-hate speech}, although the term ``câncer'' (cancer) may be found in non-pejorative contexts of use (e.g., he has cancer), in this comment context, it is used with a pejorative connotation. In contrast, the expression ``besta humana'' (human beast) and the term ``canalha'' (jerk) also present pejorative connotations even though they would be mainly found in pejorative contexts. Moving forward, it should be noted that both offensive and hate speech comments include implicit terms or expressions. For example, the expressions ``voltar a jaula'' (go back to the cage) and ``devolver o dinheiro'' (refund money) are clues that indicate the implicit pejorative terms ``criminal'' (criminoso) and ``assaltante'' (thief), respectively. Furthermore, \textit{hate speech} comments consists of attacks against groups (e.g., sexism and partyism); and \textit{non-offensive} comments do not present any terms or expressions with pejorative connotation.

Corroborating these offensive language and hate speech definitions, and taking advantage of our initial premise that a well-defined and suitable annotation schema is a great determining factor to improve the machine learning classifier's performance, we introduce in this paper a new annotation schema for hate speech and offensive language detection on social media. The proposed annotation schema is shown in Figure \ref{fig:schema}. Note that our annotation schema is divided into three layers. In the first layer, we annotated the corpus using a binary classification (offensive or non-offensive comments). Subsequently, we selected only offensive comments obtained from the previous annotation layer, and classified them into offensiveness levels. The offensiveness-level classification consists of three classes: highly, moderately, and slightly. The third layer provides annotation of offensive comments with hate speech (one of the nine hate groups that we already introduced), and offensive comments without hate speech. We further describe in detail how the classification is performed in each annotation layer to following.

\begin{itemize}
    \item \textbf{Offensive language classification}: We initially assume that comments that present at least one term or expression with any pejorative connotation should be classified as offensive, and comments that have no terms or expressions with any pejorative connotation should be classified as non-offensive comments. 
\end{itemize}

\begin{itemize}
    \item \textbf{Offensiveness-level classification}: In this paper, we introduce a fine-grained offensive annotation, which we called offensiveness-level classification. In this layer of annotation, comments classified as offensive were also annotated according to three offensiveness levels: highly, moderately, and slightly. We assume that offensive comments that present a sequence of swear words should immediately be classified as highly offensive. In the same setting, offensive comments containing a sequence of at least three terms or/and expressions with any pejorative connotation, which may be explicit or implicit, should also be classified as highly offensive. Moving forward, comments that do not meet these last two criteria, and present at least two terms or expressions with any pejorative connotation, which may be explicit or implicit, should be classified as moderately offensive. Lastly, offensive comments that do not meet the previous criteria should be classified as slightly offensive. 

\end{itemize}

\begin{itemize}
    \item \textbf{Hate speech classification}: We assume that offensive comments targeted against groups based on specific characteristics (e.g., physical appearance, religion, etc.) should be classified as hate speech. On the other hand, offensive comments not targeted against groups should not be classified as hate speech. The annotation of hate speech comments was accomplished according to nine hate speech groups (partyism, sexism, religion intolerance, apology for the dictatorship, fatphobia, homophobia, racism, antisemitism, and xenophobia).  
\end{itemize}

Therefore, the annotators followed three main steps. In the first step, they classified each of the collected Instagram comments in offensive or non-offensive comments. In the second step, for the offensiveness-level classification, each one of 3,500 comments labeled as offensive in the previous step received one of the three following labels: highly, moderately and slightly offensive. Finally, in the third step, offensive comments were classified by each annotator into nine hate speech groups. 

\subsubsection{Annotation Evaluation}
As already mentioned, our corpus was annotated by three different specialists. Each comment was annotated by each one to guarantee the reliability of the process. Besides that, the linguist and hate speech expert served as judges when a tie occurred. We also computed inter-annotator agreement using two different evaluation metrics: Cohen's kappa \cite{mchugh2012interrater,sim2005kappa} and Fleiss' kappa  \cite{fleiss1971measuring} - which we describe in detail below. A further evaluation of hate speech groups was also carried out. Firstly, the offensive comments annotated with any hate speech groups by at least two annotators were immediately validated. Then, offensive comments annotated with hate speech groups by only one annotator were submitted to a new checking step, in which the linguist decided whether that label should be validated or discarded. 

\subsubsection*{Cohen's kappa}
\label{sec:kappa}
This measure is described by equation \ref{eq:kappa}, where \textit{po} is the relative agreement observed between raters and \textit{pe} is the hypothetical probability of random agreement. It shows the degree of agreement between two or more judges beyond what would be expected by chance \cite{mchugh2012interrater,sim2005kappa}. Kappa values range from 0 to 1, and there are possible interpretations of these values. Each stratum represents the final value of the Kappa score and the level of agreement among annotators. Note that a value from 0.0 to 0.20 is a slight agreement, from 0.21 to 0.40 is fair, from 0.41 to 0.60 is moderate, from 0.61 to 0.80 is substantial, and above 0.80 refers to almost perfect agreement.

\begin{equation}
k=\frac{\rho o - \rho e}{1 - \rho e} 
\label{eq:kappa}
\end{equation}

Considering the tasks presented in this paper, we would agree that NLP highly subjective tasks encompass considerable negative impact on the inter-annotation agreement results. In other words, the more subjective the task is, the more difficult it is to obtain good inter-annotation agreement score. However, based on the obtained results shown in Table \ref{tab:kappa}, our annotation process presents substantial results according to strength of agreement from Cohen's kappa. Accordingly, high inter-annotator agreement for offensive language classification (75\%), and moderate inter-annotator agreement for offensiveness-level classification (47\%) were reached. It should be noted that, although the moderate performance obtained for the offensiveness-level classification would be a further investigation issue, we must point out that this task is highly subjective and ambiguous, consequently presenting a wide range of challenges such as high disagreement. Note that ``AB'', ``BC'', and ``CA'' consists of obtained score agreement between two different human annotators.

\begin{table}[!htb]
\caption{Cohen's kappa.}
\label{tab:kappa}
\centering
\scalefont{0.85}
\begin{tabular}{p{30mm}|p{6mm}|p{6mm}|p{6mm}|p{6mm}}
\hline
\textbf{Peer Agreement}   & \textbf{AB}   & \textbf{BC} & \textbf{CA} & \textbf{AVG}  \\
\hline
Offensive language  & 0.76 & 0.72 & 0.76 & \textbf{0.75}\\
\hline
Offensiveness-level & 0.46 & 0.44 & 0.50 & \textbf{0.47}\\
\hline
\end{tabular}
\end{table}

\subsubsection*{Fleiss' kappa}
\label{sec:fleiss}
The Fleiss evaluation measure \cite{fleiss1971measuring} is an extension of Cohen's kappa for cases where there are more than two annotators (or methods). That being said, Fleiss' kappa is applied when there is wide range of annotators that provide categorical ratings, such as binary or nominal scale, for a fixed number of items. The interpretation for the values of Fleiss' kappa also follows the values proposed by Cohen's kappa. In this paper, we also evaluated our annotation process using the Fleiss metric, as shown in Table \ref{tab:fleiss}.

\begin{table}[!htb]
\caption{Fleiss' kappa.}
\label{tab:fleiss}
\centering
\scalefont{0.85}
\begin{tabular}{p{30mm}|p{6mm}}
\hline
\textbf{Fleiss' kappa} & \textbf{ABC}\\\hline
Offensive language  &   \textbf{0.74}\\
\hline
Offensiveness-level  & \textbf{0.46}\\
\hline
\end{tabular}
\end{table}

As it is shown in Table \ref{tab:fleiss}, a high inter-annotator agreement was reached for offensive language classification (74\%), and a moderate inter-annotator agreement score (46\%) was obtained for offensiveness-levels classification. Once again, the fine-grained offensive classification is a ambitious and challenging task due to high disagreement among annotators. 

\section{HateBR Corpus Statistics}
\label{sec:hatebrstatistics}
 As a result of this paper, we present statistics of the HateBR corpus. As shown in Tables \ref{tab:binclasses} and \ref{tab:granlevels}, the corpus is composed of 7,000 document-level annotations. Firstly, the corpus was annotated into a binary class. Each of the 7,000 comments received a  offensive label (3,500 comments) or non-offensive (3,500 comments) label. Additionally, the 3,500 comments identified as offensive were also classified according to offensiveness-level, being 1,678 slightly offensive, 1,044 moderately offensive, and 778 highly offensive. As shown in Tables \ref{tab:binclasses1} and \ref{tab:granlevels1}, offensive comments were also categorized according to the nine hate speech groups (partyism, sexism, religion intolerance, apology for the dictatorship, fatphobia, homophobia, racism, antisemitism, and xenophobia). Further, over the Instagram subjects of posts, in which the comments were extracted, 70\% are related to government issues, 6,6\% fake news, 6,6\% sexism, 6,6\% racism, 6,6 \% environment, and 3,3\% economy.

\begin{table}[!htbp]
\begin{minipage}{0.5\textwidth}
\centering
\caption{Offensive language.} 
\centering
\label{tab:binclasses}
\scalefont{0.90}
\begin{tabular}{ll}
\hline
\textbf{Labels}       & \textbf{Total}  \\
\hline
Non-offensive & 3,500 \\ 
Offensive     & 3,500 \\\hline
Total & 7,000 \\
\hline
\end{tabular}
\end{minipage} 
\begin{minipage}{.5\textwidth}
 \centering
\caption{Offensiveness-level.} 
\centering
\label{tab:granlevels}
\scalefont{0.90}
\begin{tabular}{lll}
\hline
\textbf{Labels}       & \textbf{Total}  \\
\hline
Slightly offensive & 1,678  \\ 
Moderately offensive & 1,044  \\ 
Highly offensive & 778  \\\hline
Total & 3,500\\
\hline
\end{tabular}
 \end{minipage}
\end{table}

\begin{table}[!htbp]
\begin{minipage}{0.5\textwidth}
\centering
\caption{Hate speech groups.} 
\centering
\label{tab:binclasses1}
\scalefont{0.90}
\begin{tabular}{ll}
\hline
\textbf{Labels}       & \textbf{Total}   \\
\hline
Partyism &  496 \\
Sexism & 97 \\
Religion Intolerance & 47 \\ 
Apology for the Dictatorship & 32 \\
Fatphobia & 27  \\
Homophobia & 17 \\
Racism & 8  \\
Antisemitism & 2  \\
Xenophobia & 1  \\
\hline
Total & 727\\
\hline
\end{tabular}
\end{minipage} 
\begin{minipage}{.5\textwidth}
 \centering
\caption{Post subjects.} 
\centering
\label{tab:granlevels1}
\scalefont{0.90}
\begin{tabular}{ll}
\hline
\textbf{Subjects}      & \textbf{Total}  \\
\hline
Political-government  & 21 \\
Political-fake news   & 2 \\
Political-sexism      & 2 \\
Political-racism      & 2 \\
Political-environment & 2 \\
Political-economy     & 1 \\
\hline
Total & 30  \\
\hline
\end{tabular}
 \end{minipage}
\end{table}

\section{Experiments}
\label{sec:hatebr_ex}
Towards investigation and validation related to suitability of the proposed expert annotated corpus for online offensive language and hate speech detection, we implemented baseline experiments using two different representations and four machine learning methods. The representations implemented were \textit{n-grams}, more specifically the \textit{unigram language model}, and the bag-of-ngrams with tf-idf \footnote{Term Frequency(TF) — Inverse Dense Frequency(IDF)} preprocessing. The machine learning methods applied were Naive Bayes (NB) \cite{eyheramendy2003naive}, Support Vector Machine (SVM) with linear kernel \cite{scholkopf2001learning}, Multilayer Perceptron (MLP) with only one hidden layer \cite{haykin2009neural}, and Logistic Regression (LR) \cite{ayyadevara2018logistic}. In our experiments, we used the Python 3.6, scikit-learn and pandas libraries, and sliced our data in 80\% train, 10\% test, and 10\% validation. Results are shown in Table \ref{tab:eval}.

\begin{table*}[!htbp]
\scalefont{0.78} 
\caption{NB, SVM, MLP and LR Evaluation.}
\centering
\label{tab:eval}
\begin{tabular}{l|l|l|llll|llll|llll}
\hline
\multicolumn{1}{c|}{\multirow{2}{*}{Tasks}} & \multicolumn{1}{c|}{\multirow{2}{*}{Features set}} & \multicolumn{1}{c|}{\multirow{2}{*}{Class}} & \multicolumn{4}{c|}{Precision} & \multicolumn{4}{c|}{Recall} & \multicolumn{4}{c}{F1-Score} \\ \cline{4-15} 
\multicolumn{1}{c|}{} & \multicolumn{1}{c|}{} & \multicolumn{1}{c|}{} & \multicolumn{1}{c|}{NB} & \multicolumn{1}{c|}{SVM} & \multicolumn{1}{c|}{MLP} & \multicolumn{1}{c|}{LR} & \multicolumn{1}{c|}{NB} & \multicolumn{1}{c|}{SVM} & \multicolumn{1}{c|}{MLP} & \multicolumn{1}{c|}{LR} & \multicolumn{1}{c|}{NB} & \multicolumn{1}{c|}{SVM} & \multicolumn{1}{c|}{MLP} & \multicolumn{1}{c}{LR} \\ \hline
\multirow{6}{*}{\begin{tabular}[c]{@{}l@{}}Offensive \\ Language \\ Detection\end{tabular}} & \multirow{3}{*}{unigram} & 0 & \multicolumn{1}{l|}{0.72} & \multicolumn{1}{l|}{0.82} & \multicolumn{1}{l|}{0.83} & 0.83 & \multicolumn{1}{l|}{0.89} & \multicolumn{1}{l|}{0.79} & \multicolumn{1}{l|}{0.87} & 0.87 & \multicolumn{1}{l|}{0.79} & \multicolumn{1}{l|}{0.81} & \multicolumn{1}{l|}{0.85} & 0.85 \\ \cline{3-15} 
 &  & 1 & \multicolumn{1}{l|}{0.84} & \multicolumn{1}{l|}{0.78} & \multicolumn{1}{l|}{0.85} & 0.85 & \multicolumn{1}{l|}{0.62} & \multicolumn{1}{l|}{0.81} & \multicolumn{1}{l|}{0.81} & 0.81 & \multicolumn{1}{l|}{0.71} & \multicolumn{1}{l|}{0.79} & \multicolumn{1}{l|}{0.83} & 0.83 \\ \cline{3-15} 
 &  & Avg & \multicolumn{1}{l|}{0.78} & \multicolumn{1}{l|}{0.80} & \multicolumn{1}{l|}{0.84} & 0.84 & \multicolumn{1}{l|}{0.75} & \multicolumn{1}{l|}{0.80} & \multicolumn{1}{l|}{0.84} & 0.84 & \multicolumn{1}{l|}{0.75} & \multicolumn{1}{l|}{0.80} & \multicolumn{1}{l|}{0.84} & 0.84 \\ \cline{2-15} 
 & \multirow{3}{*}{tf-idf} & 0 & \multicolumn{1}{l|}{0.75} & \multicolumn{1}{l|}{0.85} & \multicolumn{1}{l|}{0.85} & 0.85 & \multicolumn{1}{l|}{0.85} & \multicolumn{1}{l|}{0.85} & \multicolumn{1}{l|}{0.85} & 0.85 & \multicolumn{1}{l|}{0.80} & \multicolumn{1}{l|}{0.85} & \multicolumn{1}{l|}{0.85} & 0.85 \\ \cline{3-15} 
 &  & 1 & \multicolumn{1}{l|}{0.81} & \multicolumn{1}{l|}{0.84} & \multicolumn{1}{l|}{0.83} & 0.83 & \multicolumn{1}{l|}{0.69} & \multicolumn{1}{l|}{0.84} & \multicolumn{1}{l|}{0.84} & 0.84 & \multicolumn{1}{l|}{0.75} & \multicolumn{1}{l|}{0.84} & \multicolumn{1}{l|}{0.84} & 0.84 \\ \cline{3-15} 
 &  & Avg & \multicolumn{1}{l|}{0.78} & \multicolumn{1}{l|}{0.85} & \multicolumn{1}{l|}{0.84} & 0.84 & \multicolumn{1}{l|}{0.77} & \multicolumn{1}{l|}{0.85} & \multicolumn{1}{l|}{0.84} & 0.84 & \multicolumn{1}{l|}{0.77} & \multicolumn{1}{l|}{\textbf{0.85}} & \multicolumn{1}{l|}{0.84} & 0.84 \\ \hline
\multirow{6}{*}{\begin{tabular}[c]{@{}l@{}} Hate Speech \\ Detection\end{tabular}} & \multirow{3}{*}{unigram} & 0 & \multicolumn{1}{l|}{0.71} & \multicolumn{1}{l|}{0.61} & \multicolumn{1}{l|}{0.69} & 0.69 & \multicolumn{1}{l|}{0.76} & \multicolumn{1}{l|}{0.79} & \multicolumn{1}{l|}{0.89} & 0.89 & \multicolumn{1}{l|}{0.73} & \multicolumn{1}{l|}{0.69} & \multicolumn{1}{l|}{0.77} & 0.77 \\ \cline{3-15} 
 &  & 1 & \multicolumn{1}{l|}{0.80} & \multicolumn{1}{l|}{0.79} & \multicolumn{1}{l|}{0.89} & 0.89 & \multicolumn{1}{l|}{0.76} & \multicolumn{1}{l|}{0.61} & \multicolumn{1}{l|}{0.68} & 0.68 & \multicolumn{1}{l|}{0.78} & \multicolumn{1}{l|}{0.69} & \multicolumn{1}{l|}{0.77} & 0.77 \\ \cline{3-15} 
 &  & Avg & \multicolumn{1}{l|}{0.76} & \multicolumn{1}{l|}{0.70} & \multicolumn{1}{l|}{0.79} & 0.79 & \multicolumn{1}{l|}{0.76} & \multicolumn{1}{l|}{0.70} & \multicolumn{1}{l|}{0.79} & 0.79 & \multicolumn{1}{l|}{0.76} & \multicolumn{1}{l|}{0.69} & \multicolumn{1}{l|}{0.77} & 0.77 \\ \cline{2-15} 
 & \multirow{3}{*}{tf-idf} & 0 & \multicolumn{1}{l|}{0.74} & \multicolumn{1}{l|}{0.64} & \multicolumn{1}{l|}{0.69} & 0.69 & \multicolumn{1}{l|}{0.77} & \multicolumn{1}{l|}{0.82} & \multicolumn{1}{l|}{0.85} & 0.85 & \multicolumn{1}{l|}{0.76} & \multicolumn{1}{l|}{0.75} & \multicolumn{1}{l|}{0.76} & 0.76 \\ \cline{3-15} 
 &  & 1 & \multicolumn{1}{l|}{0.82} & \multicolumn{1}{l|}{0.84} & \multicolumn{1}{l|}{0.86} & 0.86 & \multicolumn{1}{l|}{0.78} & \multicolumn{1}{l|}{0.71} & \multicolumn{1}{l|}{0.70} & 0.70 & \multicolumn{1}{l|}{0.80} & \multicolumn{1}{l|}{0.77} & \multicolumn{1}{l|}{0.77} & 0.77 \\ \cline{3-15} 
 &  & Avg & \multicolumn{1}{l|}{0.78} & \multicolumn{1}{l|}{0.76} & \multicolumn{1}{l|}{0.77} & 0.77 & \multicolumn{1}{l|}{0.78} & \multicolumn{1}{l|}{0.77} & \multicolumn{1}{l|}{0.78} & 0.78 & \multicolumn{1}{l|}{\textbf{0.78}} & \multicolumn{1}{l|}{0.76} & \multicolumn{1}{l|}{0.77} & 0.77 \\ \hline
\end{tabular}
\end{table*}

\begin{table*}[!htbp]
\centering
\caption{Hate speech and offensive language detection in Portuguese: datasets.} 
\scalefont{0.90}
\label{tab::compara1}
\begin{tabular}{p{25mm}|p{9mm}|p{11mm}|p{22mm}|p{35mm}|p{15mm}|p{12mm}}
\hline
 \textbf{Authors}  & \textbf{Total} & \textbf{Type} & \textbf{Classes} & \textbf{Hate Groups} & \textbf{Agreement} & \textbf{Balanced} \\\hline
 
 \newcite{fortuna2019hierarchically}  & 5,668 & tweets & hate x no-hate & sexism, body, origin, homophobia, racism, ideology, religion, health, other-lifestyle & 72\% & no  \\\hline

\newcite{MoreiraAndPelle2017} &  1,250 & website comments & offensive x non-offensive & racism, sexism, homophobia, xenophobia, religious intolerance, and cursing  & 71\% & no  \\\hline

HateBR corpus &  \textbf{7,000} &  instagram & (offensive x non-offensive); (offensiveness: slightly x moderately x highly); (hate x no-hate)  & xenophobia, racism, homophobia, sexism, religious intolerance, partyism, apology to the dictatorship, antisemitism, and fatphobia & \textbf{75\%} & yes  \\ 
\hline
\end{tabular}
\end{table*}

\begin{table*}[!htbp]
\centering
\caption{Hate speech and offensive language detection in Portuguese: models and methods.} 
\scalefont{0.92}
\label{tab:compara2}
\begin{tabular}{p{39mm}|p{30mm}|p{30mm}|p{15mm}}
\hline
\textbf{Authors}  & \textbf{Set of Features} & \textbf{Learning Method} & \textbf{F1-Score}\\ \hline
\newcite{fortuna2019hierarchically}  & Embeddings        & LSTM            & 0.78  \\\hline

\newcite{MoreiraAndPelle2017}        & N-grams, InfoGain & SVM, NB         & 0.80  \\\hline

HateBR corpus                        & N-grams, Tf-idf & NB, SVM, MLP, LR & \textbf{0.85}  \\\hline
\end{tabular}
\end{table*}

As shown in Table \ref{tab:eval}, we evaluated two different tasks: offensive language and hate speech detection. For the offensive language detection task, we implemented both baseline representations (unigram and tf-idf) over the 7,500 comments from HateBR, being 3,500 offensive labels and 3,500 non-offensive labels. As result, a high performance was achieved. The best model for this task obtained 85\% of F1-score. In the same setting, for the hate speech detection task, we also implemented both baseline representation (unigram and tf-idf) over the 3,500 offensive comments from the HateBR corpus, being 727 hate speech labels and 2,773 non hate speech labels. In this experiment, we applied a class balancing technique called undersampling \cite{witten2016data}, aiming at unbalanced classes of hate speech. In particular, we adopted this method due to the fact that it makes overfitting unlikely. As result, a high performance was also obtained for hate speech detection task. The best model obtained 78\% of F1-score. Furthermore, although the main focus of this paper is to provide a well-structured, suitable and expert annotation schema for offensive language and hate speech detection, these baseline experiments were presented to corroborate our initial premise that a well-defined and structured annotation schema leads to a good classification performance for highly complex and subjective tasks.

Despite the fact that dataset comparison is a challenging task in NLP, we propose a comparison among annotated corpora for the Portuguese language with our HateBR expert annotated corpus. Additionally, a comparison for the European and Brazilian Portuguese is also presented. Tables \ref{tab::compara1} and \ref{tab:compara2} show the results. 

Note that the proposed corpus is the first large-scale manually annotated corpus for Portuguese, which consists of 7,000 Instagram comments annotated with three different classes. Besides that, as shown in Table \ref{tab::compara1}, corpora proposed by literature for offensive language and hate speech detection in Portuguese present a considerably smaller size compared to our corpus. The inter-human agreement score obtained in HateBR also overcame the other proposals. The annotation process proposed by \newcite{MoreiraAndPelle2017}  was carried out over a corpus composed of only 1,250 comments, annotated with binary classification (offensive and non-offensive). Our corpus also presents balanced classes for offensive language classification (3,500 offensive comments  x 3,500 non-offensive comments).

Last but certainly not least, as shown in Table \ref{tab:compara2}, the results obtained with baseline experiments on our corpus clearly overcame the current ML models proposed by literature for the Portuguese language. Notice that \newcite{fortuna2019hierarchically} proposed a sophisticated set of features and ML methods, which obtained an inferior performance compared to our baseline experiments. \newcite{MoreiraAndPelle2017} also implemented models for offensive comments detection, and, even though their models have been trained over an unrepresentative corpus, the baseline experiments performed on our corpus still overcame their performance.

\section{Final Remarks}
\label{sec:final_remarks}
This paper provides the first large-scale expert annotated corpus of Brazilian Portuguese Instagram comments for offensive language and hate speech detection. The HateBR corpus was annotated by different specialists, and consists of 7,000 documents annotated with three different layers. The first layer consists of 3,500 comments annotated as offensive and 3,500 comments annotated as non-offensive. In the second layer, offensive comments were annotated according to offensiveness-level: slightly, moderately, and highly. In the third layer, offensive comments were also annotated considering nine hate speech groups. We evaluated the proposed annotation schema and a high human annotation agreement was obtained. Finally, baseline experiments were implemented, obtaining relevant results (85\% of F1-score) that overcame the current literature baselines for the Portuguese language.
\section*{Acknowledgements}
The authors are grateful to CNPq, FAPEMIG, and FAPESP for partially funding this project.

\section{Bibliographical References}\label{reference}
\bibliographystyle{lrec2022-bib}
\bibliography{lrec2022-example}


\end{document}